\documentclass[runningheads]{llncs}

\usepackage{eccv}

\usepackage{eccvabbrv}

\usepackage{graphicx}
\usepackage{booktabs}
\usepackage{enumitem}
\usepackage{wrapfig}
\usepackage{makecell}
\usepackage{booktabs}

\usepackage[accsupp]{axessibility}  

\usepackage[colorlinks=true, linkcolor=blue, urlcolor=blue, citecolor=blue]{hyperref}

\usepackage{orcidlink}

\begin{document}

\title{\LARGE MosaicMem: Hybrid Spatial Memory for Controllable Video World Models}
\titlerunning{MosaicMem}

\author{
Wei Yu$^{1,2,*,\dagger}$ \and
Runjia Qian$^{3,*}$ \and
Yumeng Li$^{*}$\thanks{Work done prior to joining BFL.} \and
Liquan Wang$^{4}$ \and
Songheng Yin$^{5}$ \and
Sri Siddarth Chakaravarthy P$^{4}$ \and
Dennis Anthony$^{4}$ \and
Yang Ye$^{6}$ \and
Yidi Li$^{3,7}$ \and
Weiwei Wan$^{3}$ \and
Animesh Garg$^{4}$
}
\authorrunning{ }
\institute{
\footnotesize
$^{1}$University of Toronto \quad
$^{2}$Vector Institute \quad
$^{3}$The University of Osaka \quad \\
$^{4}$Georgia Institute of Technology \quad
$^{5}$Mujin Inc. \quad
$^{6}$University of Texas at Austin \quad
$^{7}$Taiyuan University of Technology \\[4pt]
$^{*}$Core contributors \qquad
$^{\dagger}$Project Lead
}

\maketitle
\vspace{-2em}
\begin{abstract}
  Video diffusion models are moving beyond short, plausible clips toward world simulators that must remain consistent under camera motion, revisits, and intervention. Yet spatial memory is still a key bottleneck: explicit 3D structures can improve reprojection-based consistency but struggle with depicting moving objects, while implicit memory often produces inaccurate camera motion even with correct poses. We propose Mosaic Memory (MosaicMem), a hybrid spatial memory that lifts patches into 3D for reliable localization and targeted retrieval, while exploiting model’s native conditioning to preserve prompt-following generation. MosaicMem composes spatially aligned patches in the queried view via a patch-and-compose interface, preserving what should persist while letting the model inpaint what should evolve. With PRoPE camera conditioning and two new memory alignment methods, experiments show improved pose adherence versus implicit memory and stronger dynamic modeling than explicit baselines. MosaicMem further enables minute-level navigation,  memory-based scene editing and autoregressive rollout. For additional visual results, please check our  \href{https://mosaicmem.github.io/mosaicmem/}{\textcolor{blue}{\textbf{project page}}}.
\end{abstract}

\section{Introduction}
\label{sec:intro}
\begin{figure*}[t]
    \centering
    \includegraphics[width=1.0\textwidth]{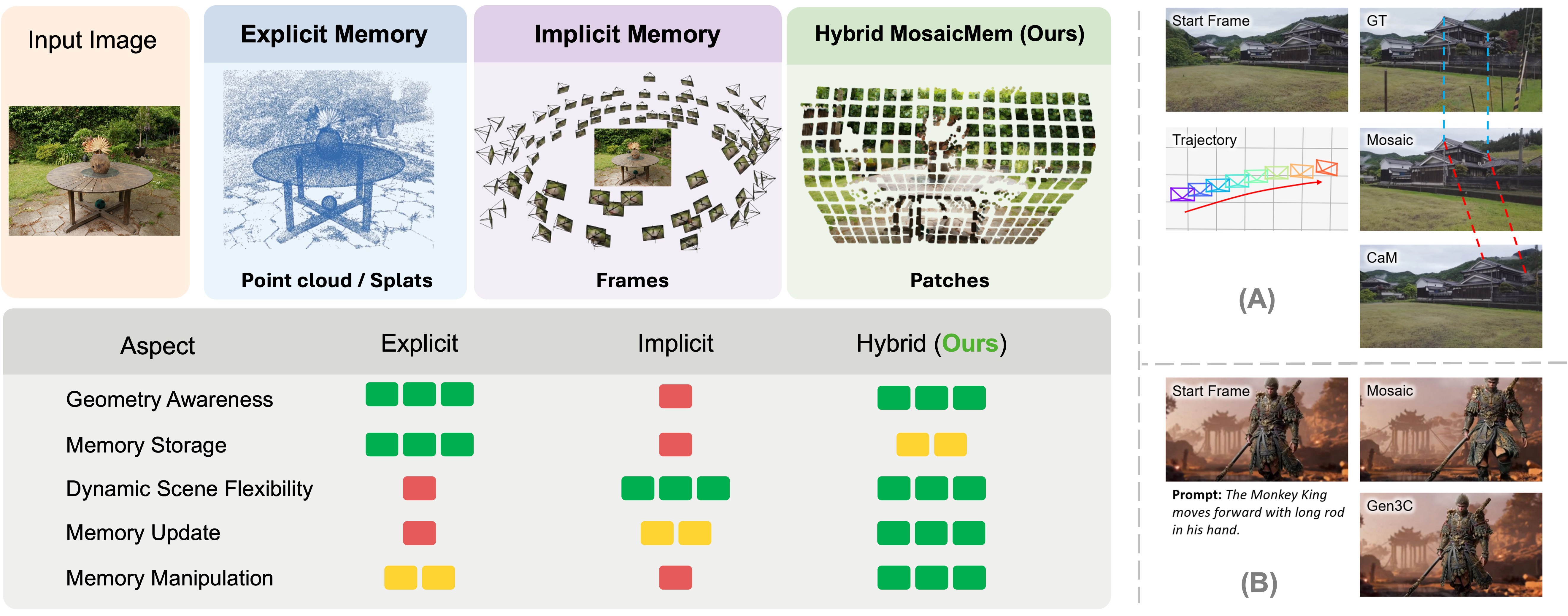}
    \caption{\textbf{Left:} Memory mechanism comparison and visualization. MoscaicMem is a hybrid approach, unifies the strengths of explicit and implicit memory. \textbf{Right:} (A) MosaicMem achieves more accurate camera motion than implicit memory.  (B) Compared to explicit memory, MosaicMem enables the generation of text-driven dynamics, whereas content generated by explicit memory remain static.}
    \label{fig:mem_comparison}
    \vspace{-1.8em}
\end{figure*}

Recent advances in video diffusion models have made high-fidelity, controllable video rollouts increasingly practical, bringing learned world simulators within reach. Such simulators \cite{parkerholder2025genie3,worldlabs2025rtfm} can empower an agent to visualize multiple plausible futures from observed environmental data—much like “playing a game” in imagination, thereby improving its ability to anticipate outcomes and respond to diverse situations. This capability reframes generative video from passive synthesis to an actionable substrate for decision-making and reinforcement learning.

The release of Genie 3 \cite{parkerholder2025genie3} illustrates what the next stage of video generation is moving toward: real-time interaction with persistence over far longer durations than typical models. The goal is no longer just plausible frames, but a coherent, explorable experience—object permanence, viewpoint consistency, and stable cause-and-effect under intervention. Achieving this kind of persistence is tightly linked to spatial memory: mechanisms that preserve and reuse scene structure across time and revisits. Yet despite rapid progress, spatial memory remains unresolved for long-horizon, physically consistent interaction; today’s designs are effective in some regimes but break in others, motivating a closer look at prevailing paradigms and their limitations. 

Broadly, spatial memory takes two forms, as visualized in Fig. \ref{fig:mem_comparison}. In explicit spatial memory \cite{ren2025gen3c,cao2025uni3c,feng2025i2vcontrol}, the system relies on external 3D estimation to build a geometric cache such as a point cloud or 3D Gaussians and, upon revisits, projects this cached structure into the queried viewpoint to condition generation, as exemplified by GEN3C \cite{ren2025gen3c}. The main advantage is that geometry is grounded by dedicated 3D inference rather than implicitly absorbed from video data, which can reduce training-data bias and improve metric faithfulness and view consistency. However, this approach fits most naturally to largely static scenes: maintaining and updating a coherent explicit cache in the presence of multiple independently moving objects remains difficult, limiting generality in dynamic environments.

Implicit memory \cite{oshima2025worldpack,yu2025contextasmem,sun2025worldplay}, by contrast, stores world state in the model’s latent representation, typically by feeding back posed frames and relying on attention for retrieval. Systems like Context-as-Memory \cite{yu2025contextasmem} and RTFM \cite{worldlabs2025rtfm} follow this approach, using spatially grounded frames as memory without building an explicit 3D scene structure. This is flexible, handling dynamics, appearance changes, and other non-rigid factors, while staying end-to-end differentiable. However, it trades off stability and efficiency: even when perfectly accurate camera poses are provided, the generated videos still exhibit inaccurate egomotion, leading to noticeable drift over revisits, and “posed-frame memory” is highly redundant, effectively converting context into memory frame-by-frame, which slows generation and caps persistence under finite context windows. Implicit state is also harder to interpret and manipulate for intricate spatial editing. Some works \cite{yu2025contextasmem,oshima2025worldpack,zhang2025packing} try to reduce context by storing compressed patch tokens, but this often degrades retrieval fidelity and long-horizon consistency, leading to blurrier or less reliable revisits.

Based on these considerations, we introduce Mosaic Memory (MosaicMem), a spatial memory design that combines the complementary strengths of both explicit and implicit paradigms. MosaicMem leverages an off-the-shelf 3D estimator to geometrically lift each patch into 3D, yielding reliable patch-level localization and recalibrated, targeted retrieval that substantially reduces the effective context required for long-term persistence. Meanwhile, it retains the advantages of implicit memory by conditioning generation through the model’s native attention mechanisms, allowing it to naturally handle dynamic, non-rigid changes. Conceptually, MosaicMem retrieves a set of spatially aligned memory patches and composes them directly in the queried view, stitching evidence onto the target frame like a mosaic that selectively fills in what must persist while leaving the model free to inpaint and update what should evolve. This structured “patch-and-compose” interface yields memory that is selective, scalable, and robust over long-horizon evolution, providing a practical path toward persistent, explorable video world simulators. Our contributions are summarized as follows:

\begin{itemize}[leftmargin=*, topsep=2pt, itemsep=2pt, parsep=0pt]
  \item 
  We propose \emph{MosaicMem}, a spatial memory mechanism that \emph{unifies explicit and implicit memory}. It leverages explicit spatial structure for precise localization and warped RoPE and latents for aligned retrieval, while exploiting model’s native conditioning to preserve prompt-following generation.

  \item 
  We incorporate \emph{PRoPE} as a principled camera conditioning interface, enabling \emph{camera-controlled video generation} with substantially improved viewpoint controllability.

  \item 
  We collect a new benchmark designed to stress-test \emph{memory retrieval under revisits}, introducing \emph{moving objects and complex camera motions} beyond the mostly static settings used in prior work.

  \item 
  Experiments show that \emph{MosaicMem} inherits complementary benefits from both paradigms: compared to implicit memory, it achieves \emph{more precise motion/pose adherence}; compared to explicit memory, it \emph{handles dynamic objects} more robustly. 
  
  \item \emph{MosaicMem} unlocks a rich set of controllable capabilities. By maintaining a long-term memory space, we demonstrate extremely long navigation video generation. The model also supports autoregressive generation. Moreover,by directly copying or relocating memory patches, we enable scene-level editing.
\end{itemize}

\vspace{-1em}
\section{Methodology}
\begin{figure*}[t]
    \centering
    \includegraphics[width=0.95\textwidth]{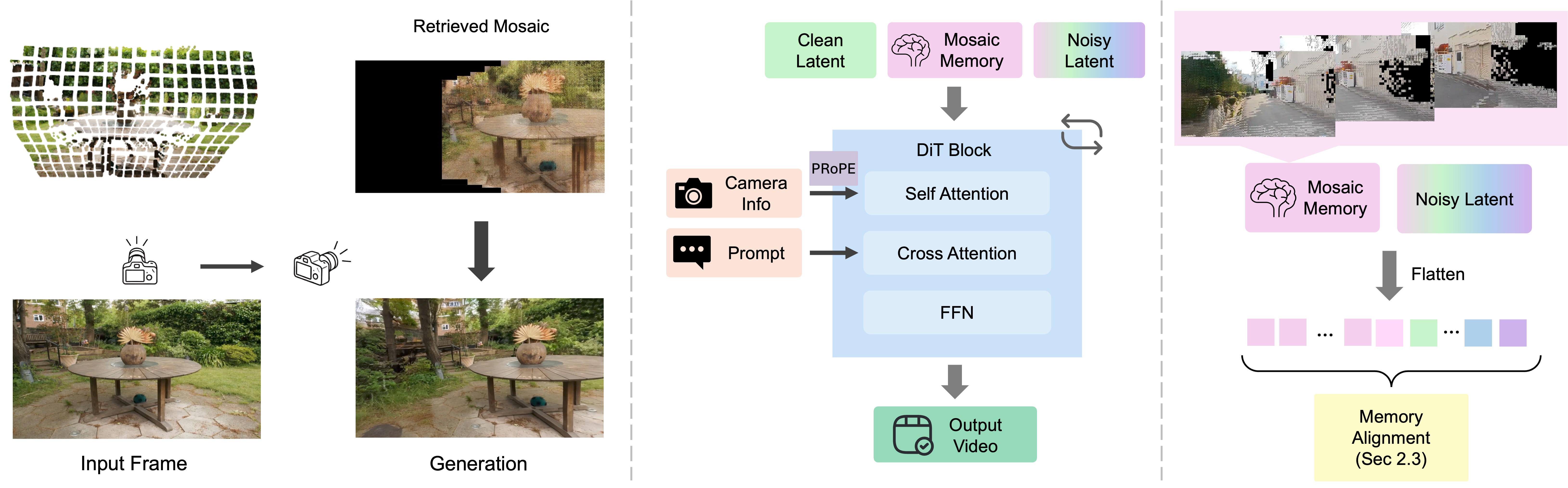}
    \vspace{-0.5em}
    \caption{\textbf{Method overview.} Left: MosaicMem lifts patches into 3D, then gathers and stitches them in the target view like a mosaic. Middle: Architecture overview. Camera motion is controlled jointly by MosaicMem retrieval and PRoPE conditioning. Right: Retrieved mosaic patches are flattened and concatenated to the token sequence as conditioning, while alignment errors were solved by warping.}
    \label{fig:method}
    \vspace{-1.8em}
\end{figure*}

\textbf{Task Definition.}
Let $\mathcal{I}$ denote  a real input image,
$\mathcal{L}=\{\ell_1,\dots,\ell_k\}$ a set of text prompts,
and $\mathcal{C}=\{{c}_1,\dots,{c}_T\}$ a sequence of camera poses.
Our goal is to generate a long-horizon video rollout
$\mathcal{X}=\{X_1,\dots,X_T\}$ that follows the specified camera trajectory,
faithfully retrieves spatial memory from real observations or previously generated clips,
and renders scene dynamics as well as unseen content consistent with the text prompts.

\noindent\textbf{Overview.}
Our method builds on text+image-to-video (TI2V) models by learning the joint distribution
of the entire video via Flow Matching.
Let $\lambda\in[0,1]$ denote the continuous flow time, and let
$\mathcal{X}^{\lambda}=\{X_1^{\lambda},\dots,X_T^{\lambda}\}$ be the video state at flow time $\lambda$.
Starting from Gaussian noise $\mathcal{X}^{0}\sim\mathcal{N}(\mathbf{0},\mathbf{I})$,
we learn a neural vector field $u_\theta$ that transports $\mathcal{X}^{0}$ to $\mathcal{X}^{1}$.
The generative process follows a probability-flow ODE:
\begin{equation}
\vspace{-0.5em}
\frac{d\mathcal{X}^{\lambda}}{d\lambda}
= u_\theta\!\left(\mathcal{X}^{\lambda}, \lambda \mid \mathcal{I}, \mathcal{L}, \mathcal{C}, \mathcal{M}\right),
\quad
 \mathcal{X}^{1}
= \mathcal{X}^{0} + \int_{0}^{1}
u_\theta\!\left(\mathcal{X}^{\lambda}, \lambda \mid \mathcal{I}, \mathcal{L}, \mathcal{C}, \mathcal{M}\right)\, d\lambda,
\end{equation}
where $\mathcal{M}$ denotes spatial memory.
Compared to the standard TI2V setting, we introduce richer conditional control,
most notably through memory retrieval and camera trajectories.
We first review the two dominant spatial memory paradigms (explicit and implicit),
highlighting their practical limitations.
Building on these insights, we introduce \textit{Mosaic Memory}, a novel design that
transcends these paradigms while combining their complementary strengths.
Furthermore, we develop an improved camera-control module tailored to modern DiT architectures.

\subsection{Preliminaries on Spatial Memory: Explicit vs.\ Implicit}

Explicit spatial memory makes the world state \emph{explicit} by \emph{lifting} information from 2D observations into an external 3D geometric cache \cite{ren2025gen3c,cao2025uni3c,feng2025i2vcontrol}, where the basic storage unit is a set of 3D primitives (e.g., points, voxels, or 3D Gaussian splats shown in Fig.~\ref{fig:mem_comparison}) rather than images. Upon revisits, memory retrieval is \emph{optics-based}: the cached 3D structure is \emph{projected} or \emph{rendered} into the queried viewpoint to produce view-aligned conditioning signals, which are typically injected into the generator through mechanisms such as ControlNet-style branches \cite{wu2025video} or channel concatenation \cite{ren2025gen3c}. The downstream video generator therefore behaves largely as \emph{video inpainting}, filling uncertain or unseen regions while being anchored by projected, geometry-consistent evidence. While this paradigm directly enforces geometric consistency, it restricts generative flexibility and rarely produces rich text-driven dynamics. Furthermore, since explicit memory is maintained through global 3D reconstruction, small cross-view misalignments accumulate over time, introducing artifacts and making long-horizon memory updates brittle.

By contrast, \emph{implicit} spatial memory performs \emph{no lift} into an explicit 3D space \cite{oshima2025worldpack,yu2025contextasmem,sun2025worldplay}. Memory remains as posed frames (or frame-derived features), with the \emph{frame} as the basic storage unit. Retrieval is mediated through the DiT’s built-in (or augmented) \emph{conditioning} mechanisms, typically via token concatenation, which select and inject relevant reference regions into generation. 
\begin{wrapfigure}{l}{0.50\linewidth}
\vspace{-1.5em}
    \centering
    \includegraphics[width=\linewidth]{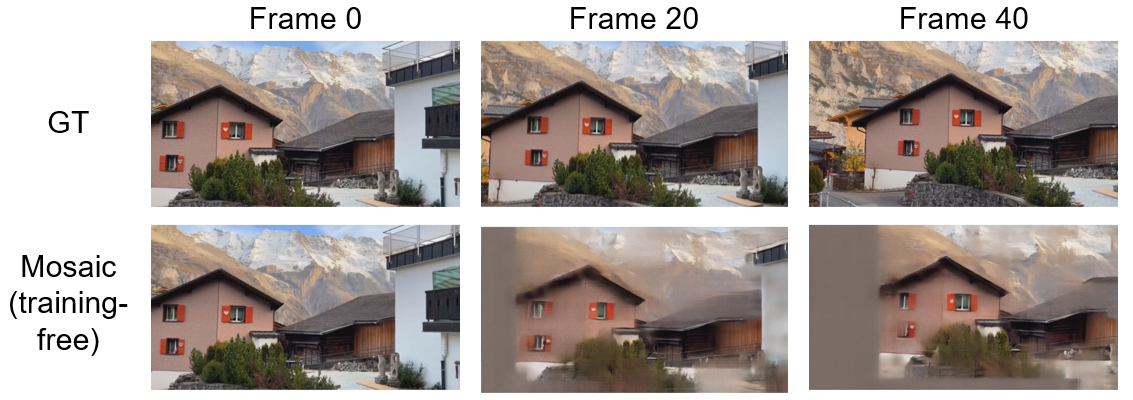}
    \vspace{-1.5em}
    \caption{Training-free generation via direct Mosaic Memory injection. 
    Without fine-tuning, the model places retrieved memory conditions at the targeted locations 
    and modestly refines it.}
    \label{fig:training_free}
\vspace{-1.8em}
\end{wrapfigure} 
This design directly exploits the model’s \emph{prompt-following capability} acquired during large-scale pretraining, allowing retrieved frames to serve as conditioning signals while naturally accommodating dynamic entities and non-rigid changes without committing to a fixed geometric parameterization. However, because viewpoint transfer is not enforced by projection, small pose errors can accumulate into spatial drift across revisits; and although recent works attempt to increase efficiency via hierarchical compression \cite{zhang2025packing,oshima2025worldpack}, the underlying frame-based representation remains highly redundant, stressing both speed and finite context windows. Moreover, since memory is stored as posed frames rather than an explicit structure, it cannot be directly manipulated through geometric operations, which motivates hybrid designs that combine the manipulability and view consistency of explicit structure with the adaptability of implicit retrieval.

\subsection{Mosaic Memory}
As discussed earlier, explicit and implicit memory differ in their fundamental memory units: explicit methods store scene evidence as points or splats, whereas implicit methods retain memory at the granularity of entire video frames. We observe an intermediate representation—\textbf{patches}—that has been  unexplored in prior work. Motivated by this observation, we propose Mosaic Memory, a new spatial memory mechanism that uses  patches as the basic unit of memory and integrates the complementary strengths of both explicit and implicit memory.

More concretely, for a given patch \(P\), we first perform a geometric lifting step---analogous to the \emph{front half} of an explicit-memory pipeline.  We use an off-the-shelf 3D estimator to infer depth together with the associated camera information, and lift the patch into 3D. 


When the observer moves to a new viewpoint and later revisits patch \(P\), we adopt a conditioning strategy analogous to the \emph{back half} of an implicit-memory pipeline: the retrieved patch is provided to the DiT as context, and a modified RoPE mechanism (\S\ref{sec:alignment}) conveys the correspondence between this memory patch and the noised latent tokens under the queried camera. In this stage, an additional camera control module (\S\ref{sec:prope}) provides fine-grained intra-patch motion guidance on the current viewpoint, enabling the model to better align retrieved spatial memory with the queried camera. The generator can flexibly decide whether to rely on spatial memory for consistent reconstruction or to synthesize unseen content and new dynamics according to the text prompt.
This organic combination---explicit-style lifting followed by implicit-style conditioning---resembles how a mosaic is assembled, stitching localized pieces into a coherent whole. We therefore call this hybrid memory mechanism \emph{Mosaic Memory}, (Fig. \ref{fig:method}).

To validate that this pipeline works, we integrated Mosaic Memory directly into the vanilla Wan 2.2 \cite{wan2025wan} without any modifications. Surprisingly, as shown in Fig. \ref{fig:training_free}, even without any additional training, the model can still project the Mosaic Memory provided in the context conditions to the correct spatiotemporal locations and generate meaningful visual content.

\textbf{Merits}: The most obvious advantage of Mosaic Memory is that it combines the strengths of the two existing paradigms. On one hand, it leverages off-the-shelf 3D estimators, as used in explicit memory, enabling more accurate camera-motion alignment and more geometrically consistent 3D scene evolution. On the other hand, it adopts the conditioning mechanism of implicit memory: the retrieved Mosaic Memory serves only as a reference signal, allowing the model to  decide whether to rely on spatial memory or to generate new text-driven dynamics. The corresponding results are presented in the evaluation section.

Mosaic Memory also introduces several appealing properties. (1) \textbf{Flexible retrieval}. Retrieval can be either dense or sparse: due to the high redundancy of video, distributing memory from the same scene across different spatiotemporal locations often suffices to reconstruct the entire sequence. Moreover, since modern video generators naturally preserve details from the first frame, regions overlapping with the initial frame do not require complete Mosaic Memory to be supplied, substantially reducing the number of conditioning tokens and alleviating a key limitation of implicit memory. (2) \textbf{Manipulable memory space}. Mosaic Memory provides a deletable and manipulable memory space in which individual object patches can be explicitly displaced, duplicated, or removed, enabling direct manipulation of spatial memory. (3) \textbf{Robust long-horizon updates}. Because Mosaic Memory stores independent localized patches rather than maintaining a globally reconstructed structure, it avoids the accumulation of cross-view misalignment, leading to more stable memory updates over extended horizons.


\subsection{Memory Alignment Through Warping}\label{sec:alignment}


While Mosaic Memory is promising, the high spatiotemporal compression of 3D VAEs introduces spatial-temporal ambiguity and reduces the effective RoPE coordinate resolution in DiT. As a result, retrieved patches may not align with the exact center of the generated region, and the limited coordinate precision can degrade reprojection accuracy, leading to local geometric inconsistencies or blurred details. To establish geometry-consistent correspondence between retrieved memory patches and the current view, we improve alignment using two warping mechanisms: warped RoPE and warped latent.

\textbf{Warped RoPE} is a new positional encoding mechanism that aligns patches across time and camera motion in latent space, driven by pixel-accurate correspondences. 
Each retrieved memory patch $P$ is associated with depth $D$ and camera intrinsics/extrinsics $(\mathbf{K}_i, \mathbf{T}_i)$ at its source timestep. Given its original RoPE coordinates $(u,v)$, we back-project the patch into 3D world space and re-project it into the target camera $(\mathbf{K}_j, \mathbf{T}_j)$ at time $j$,
\vspace{-0.1em}
\begin{equation}
(u',v') = \Pi\!\left(\mathbf{K}_j \mathbf{T}_j \mathbf{T}_i^{-1} \mathbf{K}_i^{-1} (u,v,D)\right),
\label{eq:video_dist}
\end{equation}
\vspace{-0.2em}
where $\Pi(\cdot)$ denotes the perspective projection that converts homogeneous coordinates to image-plane coordinates via perspective division.

The tuple $(j, u', v')$ jointly defines the 3D RoPE coordinate associated with this patch $P$. We preserved the fractional part of the reprojected coordinate and sampled RoPE at a higher resolution to retain as much accuracy as possible. 

Alternatively, \textbf{Warped Latent} offers a complementary alignment mechanism by directly transforming the retrieved memory patches in the feature space, rather than modifying the positional encodings. Utilizing the dense geometric correspondence established by the reprojected coordinates $(u',v')$ in Eq.~\eqref{eq:video_dist}, we perform spatial resampling on the source latent representations.

Specifically, the warped latent patch is obtained by applying differentiable bilinear grid sampling to the original latent features at the fractional coordinates $(u',v')$. These two warping mechanisms exhibit complementary advantages, particularly under autoregressive generation. 
Empirically, we find that training with a mixture of both warping strategies yields the best performance.


\subsection{PRoPE for Camera Control}\label{sec:prope}
Although Mosaic Memory implicitly provides some camera-motion cues, it is insufficient for reliable trajectory control. 
We therefore introduce a dedicated \textbf{camera control module} for 3 reasons: 
\textbf{(a)} Under large camera motions or sparse memory settings, Mosaic Memory mainly acts as a source of visual cues rather than a precise motion signal, making explicit trajectory specification necessary; 
\textbf{(b)} due to the $4\times$ temporal compression of the 3D VAE, Mosaic Memory does not capture fine-grained inter-frame motion, which is compensated by explicitly injecting frame-level motion through the camera module; 
\textbf{(c)} adding camera control enables direct reuse of previously generated video latents for faster generation without re-encoding. Although these latents already encode inter-frame motion that would otherwise propagate to the new prediction, the camera control module corrects and realigns such motion with the desired trajectory.

In this paper, we adopt Projective Positional Encoding (PRoPE) \cite{li2025prope} as a principled camera-conditioning interface for DiT-based video generation by injecting relative camera frustum geometry directly into self-attention. Given per-frame camera projection matrices $\tilde{\mathbf P}_i\in\mathbb{R}^{4\times 4}$, PRoPE encodes the complete relative relationship between two views via the projective transform $\tilde{\mathbf P}_{i_1}\tilde{\mathbf P}_{i_2}^{-1}$, and applies it through GTA-style transformed attention: $\mathrm{Attn}_{\text{PRoPE}}(\mathbf Q,\mathbf K,\mathbf V)=\mathbf D\odot \mathrm{Attn}\!\big(\mathbf D^{\top}\odot \mathbf Q,\ \mathbf D^{-1}\odot \mathbf K,\ \mathbf D^{-1}\odot \mathbf V\big)$, where each token $t$ uses a block-diagonal matrix $\mathbf D^{\text{PRoPE}}_{t}=\begin{bmatrix}\mathbf D^{\text{Proj}}_{t}&0\\0&\mathbf D^{\text{RoPE}}_{t}\end{bmatrix}$ with $\mathbf D^{\text{Proj}}_{t}=\mathbf I_{d/8}\otimes \tilde{\mathbf P}_{i(t)}$ and $\mathbf D^{\text{RoPE}}_{t}$ providing the usual 2D patch RoPE terms. The key difference in video generation is temporal compression: our spatio-temporal tokens are produced from a VAE that compresses time by a factor $s=4$, so one latent frame index $\ell$ corresponds to four original frames $\{4\ell+k\}_{k=0}^{3}$, i.e., a single latent slice must be conditioned on four camera matrices $\{\tilde{\mathbf P}_{\ell,k}\}_{k=0}^{3}$. Concretely, instead of using a single $\tilde{\mathbf P}_{i(t)}$ per token as in frame-to-frame NVS, we “unfold” an extra sub-index $k$ and apply $\mathbf D^{\text{Proj}}_{\ell,k}=\mathbf I_{d/8}\otimes \tilde{\mathbf P}_{\ell,k}$ (equivalently, pack cameras as $\tilde{\mathbf P}\in\mathbb{R}^{B\times L\times 4\times 4\times 4}$ and broadcast the $k$-indexed transforms into the Q/K/V rotations), ensuring each temporally-compressed latent frame attends with the correct per-frame projective conditioning while keeping the PRoPE interface unchanged at the attention operator level.

\section{Data Curation}

We present a new benchmark called MosaicMem-World to support training and evaluation with a particular focus on spatial memory under viewpoint changes. We observe that most publicly available first-person video datasets\cite{grauman2022ego4d,ling2024dl3dv,wang2025egovid} are dominated by forward navigation, where explicit revisitation is rare and long-range returns to previously observed areas are underrepresented. This is a poor match for evaluating whether a model can (i) retain stable scene structure over time, (ii) leave and later re-localize under substantial camera motion, and (iii) reuse stored geometry and semantics instead of re-synthesizing them. To address this gap, we intentionally collect trajectories that periodically revisit earlier checkpoints and regions within the same episode, spanning both short and extended time horizons. MosaicMem-World aggregates data from four complementary sources, each contributing on the order of tens of hours: (1) curated Unreal Engine 5 scenes built from licensed assets, where we record trajectories with single and mixed actions as well as explicit revisited segments, enabling decoupled control, flexible action composition, and long-range memory retrieval; (2) commercial game environments, e.g., Cyberpunk 2077 \cite{cyberpunk2077}, to capture dense interaction opportunities and complex world dynamics; (3) real-world first-person captures to introduce realistic appearance, noise, and illumination variations; and (4) existing datasets such as Sekai \cite{li2025sekai}, from which we select sequences with the highest revisit frequency according to the provided camera trajectories.

To standardize supervision across sources, we adopt a unified preprocessing and annotation pipeline. For each video, we reconstruct depth and camera motion using Depth Anything V3 \cite{depthanything3} or VIPE \cite{huang2025vipe}, providing a consistent geometric scaffold for learning view-consistent representations and for evaluating scene recall under revisitation. We then annotate videos in a segment-wise manner: each sequence is partitioned into fixed-length segments of 32 frames, and Gemini 3 \cite{gemini3google} generates two complementary textual descriptions per segment—one describing the static scene content in the first frame (layout, salient objects, and spatial relations), and another describing the temporal dynamics over the remaining frames (camera motion, interactions, and state changes). This factorized “static + dynamic” labeling supports compositional training: by concatenating dynamic descriptions across consecutive segments, we can construct training clips of arbitrary length. Finally, we filter the dataset by removing videos with inaccurate 3D estimates or excessive motion blur.

\begin{figure}[t]
    \centering
    \includegraphics[width=0.95\linewidth]{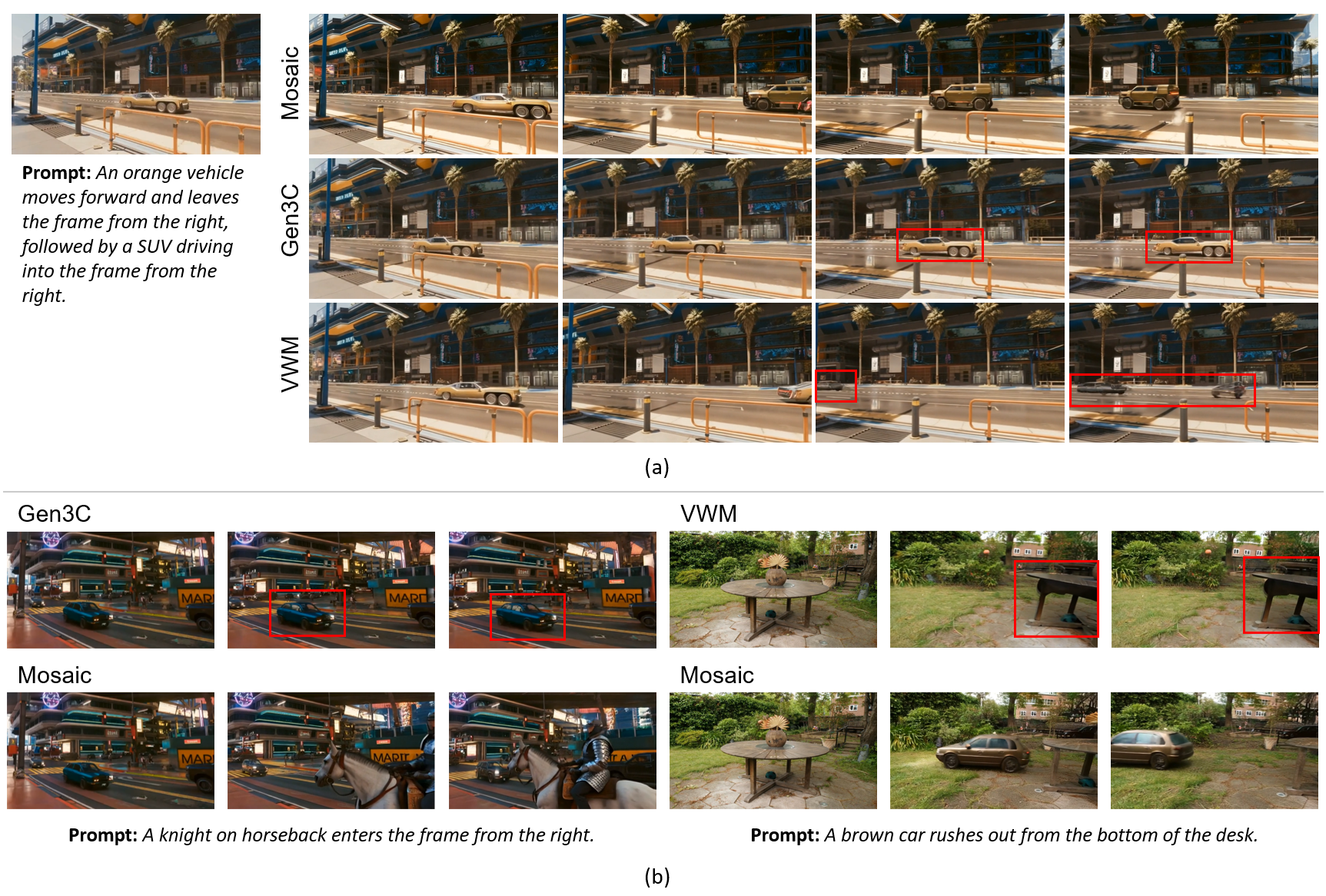}
    \vspace{-1em}
    \caption{(a) MosaicMem generates dynamic objects with temporal consistency, while GEN3C produces static scenes and VWM introduces artifacts despite supporting limited dynamics. (b) MosaicMem preserves prompt adherence to create composite scenes.}
    \label{fig:dynamic}
    \vspace{-1.6em}
\end{figure}

\section{Evaluation}

\textbf{Overview}
We first describe the training setup for fine-tuning Wan 2.2  with MosaicMem, then evaluate spatial memory (\S\ref{sec:spatial-memory}) and ablate memory alignment (\S\ref{sec:ablation}). We further demonstrate minute-level navigation with persistent memory (\S\ref{sec:long-horizon}), and memory manipulation for scene editing (\S\ref{sec:memory-manip}). Finally, we explore autoregressive video generation for efficient long-horizon synthesis (\S\ref{sec:autoregressive}).

\noindent\textbf{Implementation details.}
We fine-tune {Wan} 2.2 5B \cite{wan2025wan}, an 
open-sourced TI2V DiT, using the AdamW optimizer with a learning rate of $1\times 10^{-5}$  . We train for $250$k steps with an effective batch size of $64$ on 8 $\times$ H100 GPU clusters.
At inference time, we follow Wan's default sampler with $50$ denoising steps. 

\noindent\textbf{Metrics}: We evaluate video quality, camera control, motion dynamics, and memory retrieval using four complementary metric groups, each with a dedicated evaluation set: (1) FID\cite{bynagari2019gans} and FVD\cite{unterthiner2019fvd} for overall generation quality; (2) RotErr/TransErr for camera motion accuracy, with poses estimated by DA V3 \cite{depthanything3}; (3) Dynamic Score (Average Optical Flow Magnitude) for motion intensity, computed on trajectories with minimal camera motion to emphasize object dynamics; and (4) Consistency Score, our metric for memory retrieval accuracy, measuring SSIM\cite{wang2004image}, PSNR and LPIPS\cite{zhang2018unreasonable} within annotated corresponding regions between input and future frames.

\begin{table}[t]
\centering
\scriptsize
\setlength{\tabcolsep}{1pt}
\renewcommand{\arraystretch}{1.0}

\begin{tabular}{lccccccc c}
\toprule

\textbf{Method} &
\multicolumn{2}{c}{\textbf{Camera Control}} &
\multicolumn{2}{c}{\textbf{Visual Quality}} &
\multicolumn{3}{c}{\textbf{Consistency Score}} &
\textbf{Dynamic} \\

\cmidrule(lr){2-3} \cmidrule(lr){4-5} \cmidrule(lr){6-8}

& RotErr ($^\circ$)$\downarrow$ & TransErr$\downarrow$
& FID$\downarrow$ & FVD$\downarrow$
& SSIM$\uparrow$ & PSNR$\uparrow$ & LPIPS$\downarrow$
& $\uparrow$ \\

\midrule
\multicolumn{9}{l}{\textbf{Explicit Memory}} \\

VMem \cite{li2025vmem} & 1.59 & 0.14 & 77.12 & 363.34 & 0.64 & 21.64 & 0.17 & 1.18 \\
GEN3C \cite{ren2025gen3c} & 1.61  & 0.13 & 77.41 & 372.08 & 0.64 & 21.58 & 0.17 & 1.21 \\
SEVA \cite{zhou2025stable}& 1.42  & 0.12 & 74.67 & 301.77 & 0.66 & 22.01 & 0.15 & 1.22 \\
VWM \cite{wu2025video}  & 1.50  & 0.13 & 75.83 & 323.67 & 0.65 & 21.86 & 0.16 & 1.41 \\

\midrule
\multicolumn{9}{l}{\textbf{Implicit Memory }} \\

WorldMem \cite{xiao2025worldmem}  & 5.87 & 0.49 & 85.72 & 403.50 & 0.47 & 15.34 & 0.46 & 1.67 \\
CaM \cite{yu2025contextasmem}  & 4.65  & 0.43 & 85.32 & 392.11 & 0.49 & 15.78 & 0.42 & 1.72 \\

\midrule
\multicolumn{9}{l}{\textbf{Ablations}} \\

ControlMLP alone & 6.51 & 0.52 & 89.17 & 458.45 & 0.37 & 13.55 & 0.56 & 1.84 \\
PRoPE alone  & 4.91 & 0.36 & 86.44 & 412.85 & 0.45 & 14.32 & 0.52 & 1.75 \\
MosaicMem w/o PRoPE  & 0.79 & 0.11 & 73.18 & 250.84 & 0.68 & 22.33 & 0.14 & 2.11 \\
PRoPE + Warped Latent  & 0.66 & 0.08 & 75.46 & 268.13 & 0.65 & 21.49 & 0.15 & 1.98 \\
PRoPE + Warped RoPE  & 0.70 & 0.09 & 71.89 & 243.59 & 0.69 & 22.80 & 0.12 & 2.24 \\
\midrule
MosaicMem (full)        & \textbf{0.51} & \textbf{0.06} & \textbf{65.67} & \textbf{232.95} & \textbf{0.75} & \textbf{23.57} & \textbf{0.11} & \textbf{2.58} \\
\bottomrule
\end{tabular}
\vspace{+0.5em}
\caption{Quantitative comparison across explicit memory, implicit memory, and MosaicMem variants.
MosaicMem achieves the best performance across camera control, visual quality, retrieval consistency, and motion dynamics.}
\vspace{-2.5em}
\label{tab:memory_comparison}
\end{table}





\subsection{Spatial Memory}\label{sec:spatial-memory}
\textbf{MosaicMem v.s. Explicit Memory.} We first compare MosaicMem against explicit-memory baselines, GEN3C \cite{ren2025gen3c}, SEVA \cite{zhou2025stable}, Vmem \cite{li2025vmem} and VWM \cite{wu2025video}. The quantitative results are reported in the upper part of Table~\ref{tab:memory_comparison}. Explicit-memory baselines achieve a non-zero Dynamic Score due to limited camera motion; however, their motion dynamics remain substantially weaker than those of MosaicMem. Because explicit-memory approaches lack dynamic scene evolution, their generated distribution deviates from the true data distribution, which also leads to degraded visual quality. Moreover, our method achieves more accurate camera motion, partially benefiting from increasingly powerful DA V3 \cite{depthanything3}.

Qualitative results are shown in Figure~\ref{fig:dynamic}. The baseline models fail to generate novel moving objects and are unable to animate objects that have already been observed. In contrast, MosaicMem can follow the prompt to generate compositional scenes—for example, producing a medieval knight riding a horse in a sci-fi environment, or a car driving through a garden.

\begin{figure*}[t]
    \centering    \includegraphics[width=0.95\textwidth]{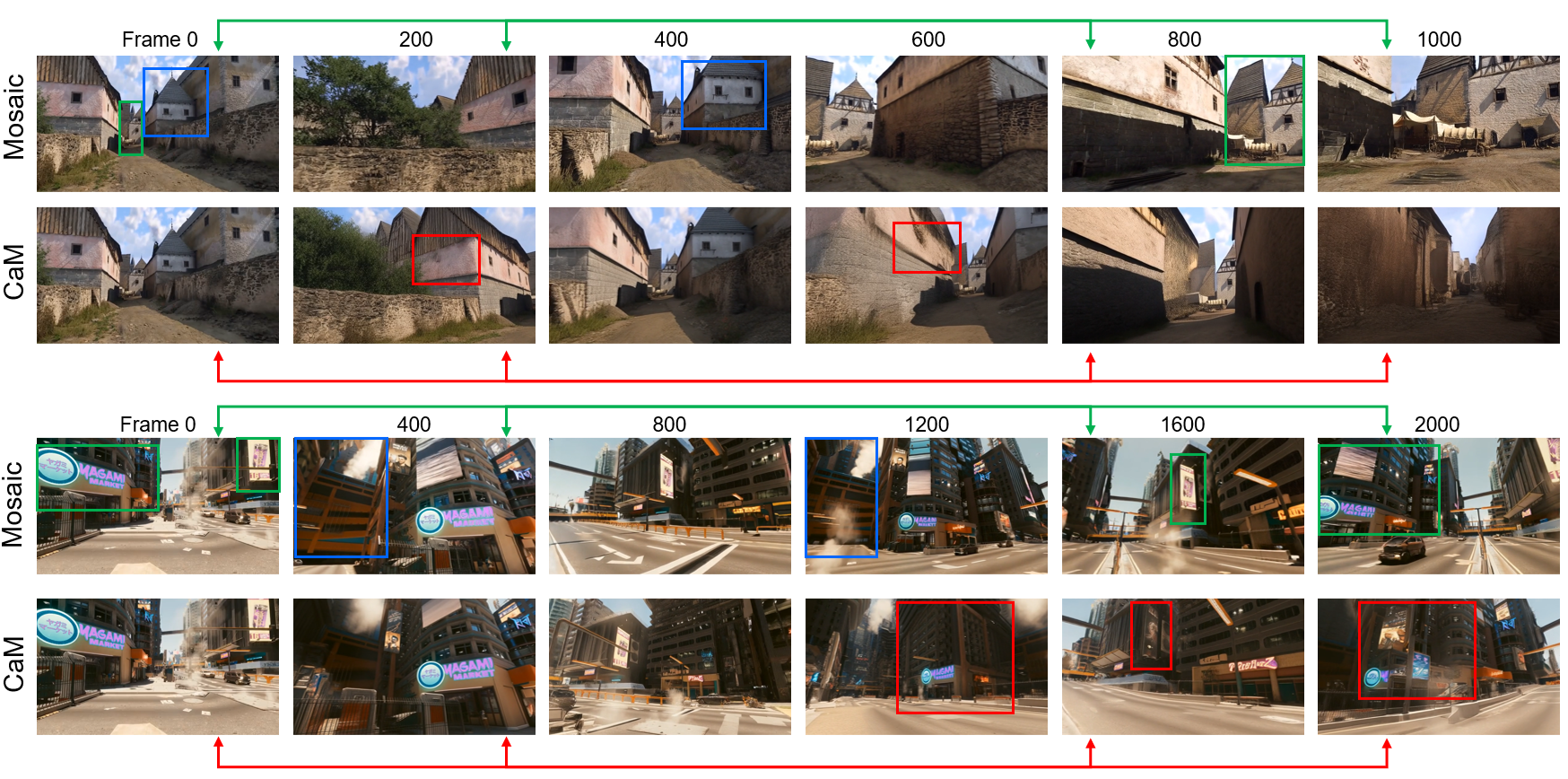}
    \caption{Through robust MosaicMem, we can generate minute-level videos. The blue and green boxes highlight that MosaicMem maintains strong consistency even after long-duration and large-scale camera motions, while the red boxes indicate artifacts and inconsistencies produced by CaM over extended sequences.}
    \label{fig:long}
    \vspace{-0.8em}
\end{figure*}

\noindent \textbf{MosaicMem v.s. Implicit Memory}: We then compare MosaicMem against implicit-
memory baselines including WorldMem\cite{xiao2025worldmem} and Context-as-Memory(CaM)\cite{yu2025contextasmem}. The quantitative results are summarized in the middle of Table~\ref{tab:memory_comparison}. MosaicMem significantly outperforms the baselines in both camera motion accuracy and memory retrieval. Its video quality is also higher due to a more faithful modeling of the camera motion distribution. The visualizations in Fig.~\ref{fig:implicit} (a)  and ~\ref{fig:long}  further corroborate these quantitative gains: even over long horizons, MosaicMem exhibits negligible camera drift and faithfully reconstructs previously observed scenes.


\subsection{Ablation Study}\label{sec:ablation}
In this section, we analyze the contributions of each component to camera motion accuracy and memory retrieval. As a baseline, we adopt ControlMLP \cite{kant2025pippo} for camera control and compare it with several alternatives: PRoPE alone, MosaicMem alone, the two individual warping mechanisms in Mosaic Memory, and the full model that combines both warping mechanisms within the memory module, which improves overall robustness.

\begin{figure*}[t]
    \centering
    \includegraphics[width=0.95\linewidth]{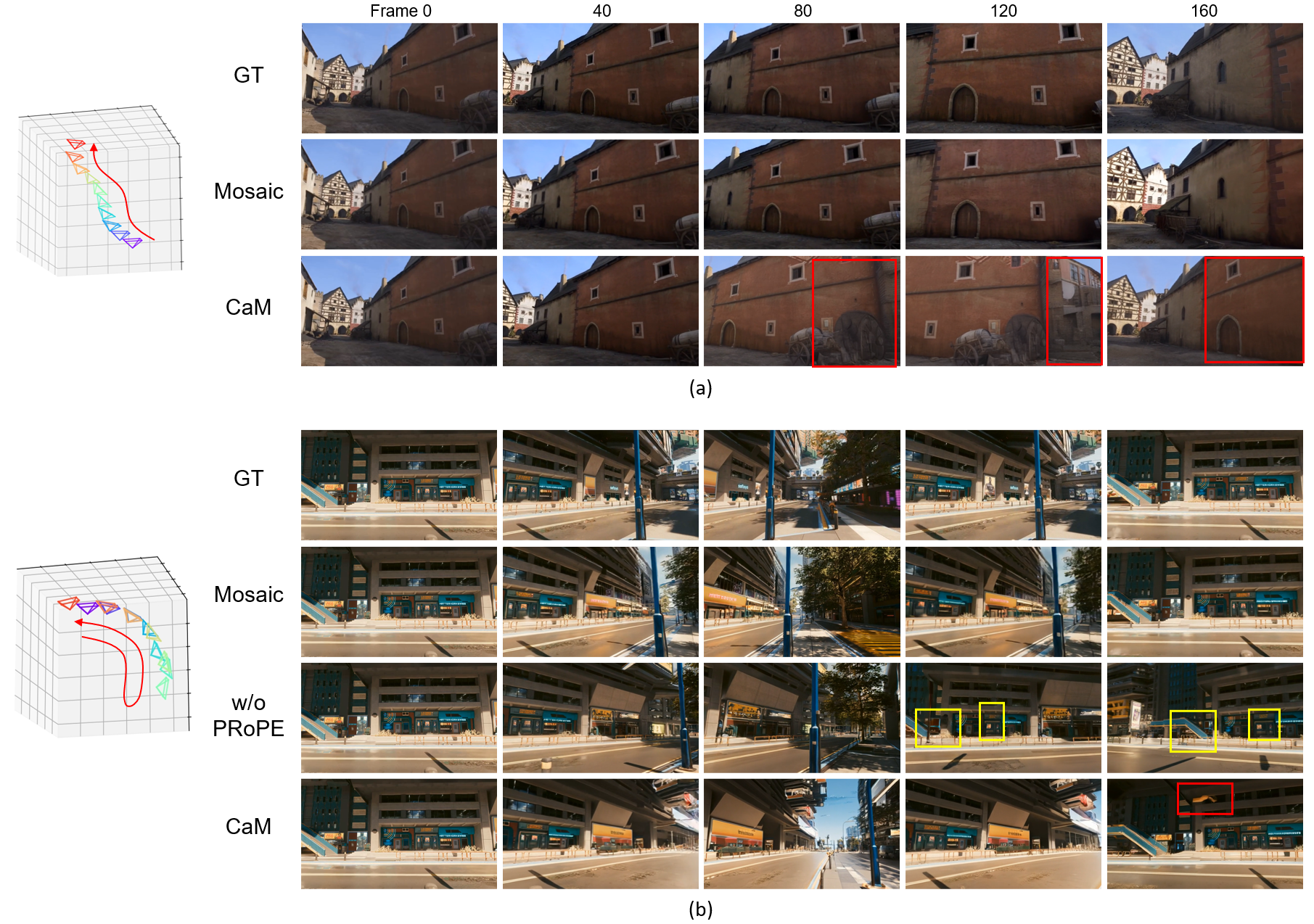}
    \vspace{-0.5em}
    \caption{(a) Comparison of camera-controlled generation between the implicit memory baseline and MosaicMem. Combining PRoPE with MosaicMem improves camera motion control and enables precise spatial memory registration.  (b) Without PRoPE, MosiacMem alone struggles with large rotation, where the camera enter previously unseen regions without Mosaic references, leading to significant camera errors.}
    \label{fig:implicit}
    \vspace{-2.2em}
\end{figure*}

The quantitative results are summarized in the bottom of Table \ref{tab:memory_comparison}. We first compare ControlMLP, PRoPE, and MosaicMem alone. The results show a progressive improvement in both camera motion accuracy and memory retrieval performance: PRoPE outperforms ControlMLP, and MosaicMem alone further improves over PRoPE. However, MosaicMem alone remains significantly weaker than the full MosaicMem model. This is because MosaicMem alone cannot properly handle fine-grained intra-patch motion. Moreover, under large camera movements, Mosaic Memory may fail to retrieve enough patches, leaving the model without effective guidance for camera-controlled generation. This observation is also confirmed by the visualization in Figure \ref{fig:implicit}(b).
 
In addition, we compare the two warping mechanisms. We observe that Warped Latent achieves more accurate camera motion overall, but its generated visual quality and memory retrieval performance are inferior to those of Warped RoPE. In subsequent autoregressive experiments, we further observed that Warped RoPE introduces another issue: objects newly observed in previous frames tend to be repeatedly regenerated near the image boundaries. By training with a mixture of both warping strategies, we obtain the final version of our model, which leads to more robust memory conditioning.

\subsection{Long-horizon Video Generation}\label{sec:long-horizon}
\textbf{Setup.} A primary role of spatial memory is to enable consistent alignment between past observations and newly generated content during long-horizon video generation. Accordingly, we conduct long-video generation experiments in this setting. Specifically, each time we generate an 80-frame segment, it is incorporated into the memory space for updating. The last frame of the previous segment is then used as the first frame of the next segment, and this process is repeated to produce a 2-minute navigation video.

\noindent\textbf{Results.} The long-term video generation results are shown in Figure \ref{fig:long}, where we use CaM as a baseline to highlight the advantages of our model. We observe that our approach not only accurately recalls previously observed regions but also faithfully renders moving objects. In contrast, CaM still suffers from inaccurate camera motion; in long video rollouts, artifacts progressively accumulate and eventually cause its generation to collapse.

\vspace{-0.5em}
\subsection{Memory Manipulation}\label{sec:memory-manip}
\vspace{-0.5em}
\textbf{Setup.} MosaicMem stores patches with their underlying 3D spatiotemporal locations. We can directly manipulate object placement by editing these locations, enabling operations such as deletion, duplication, concatenation, and relocation. Here, we explore the simplest implementation of this capability: concatenating scenes. By registering videos or images from two different scenes into Mosaic memory and spatially connecting these scene memories, we can construct a unified scene representation. This enables the generation of scenes that differ in visual style while remaining geometrically continuous, allowing users to seamlessly move between and explore the connected scenes.

\begin{figure}[t]
    \centering
\includegraphics[width=0.95\linewidth]{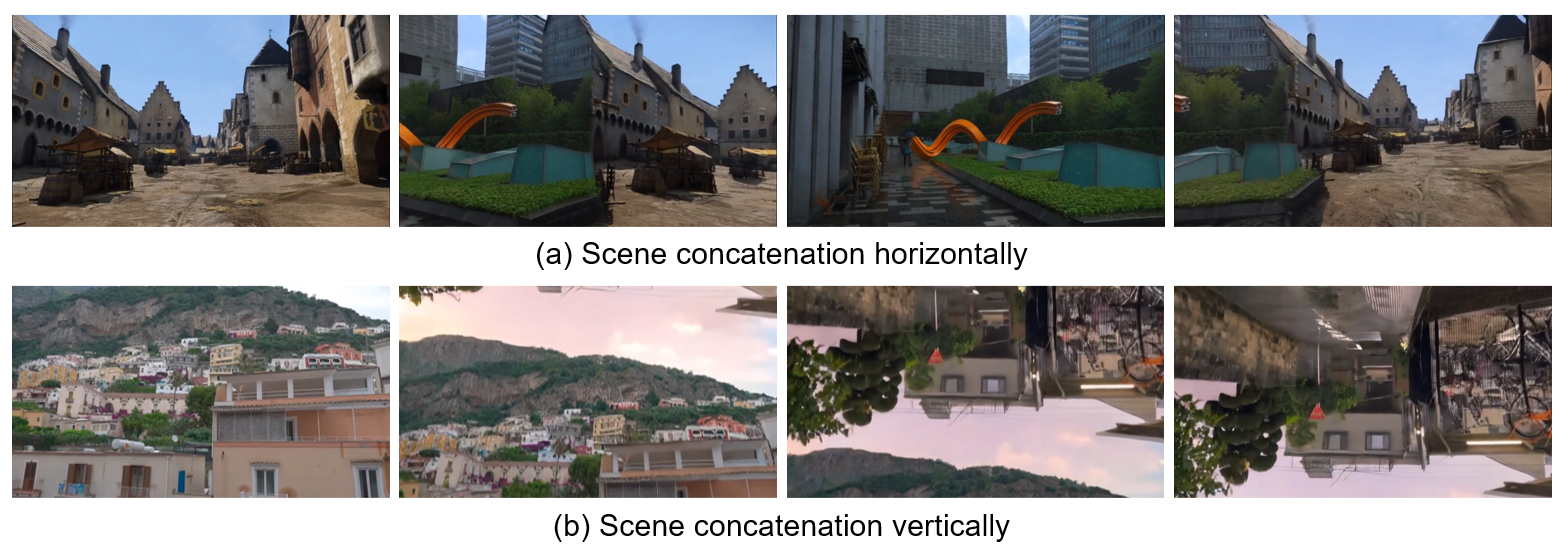}
    \vspace{-0.8em}
    \caption{Two examples of manipulating MosaicMem by stitching together two distinct scenes, along the horizontal and vertical directions, respectively.}
    \label{fig:manipulation}
    \vspace{-0.2em}
\end{figure}

\begin{figure}[t]
    \centering
\includegraphics[width=0.95\linewidth]{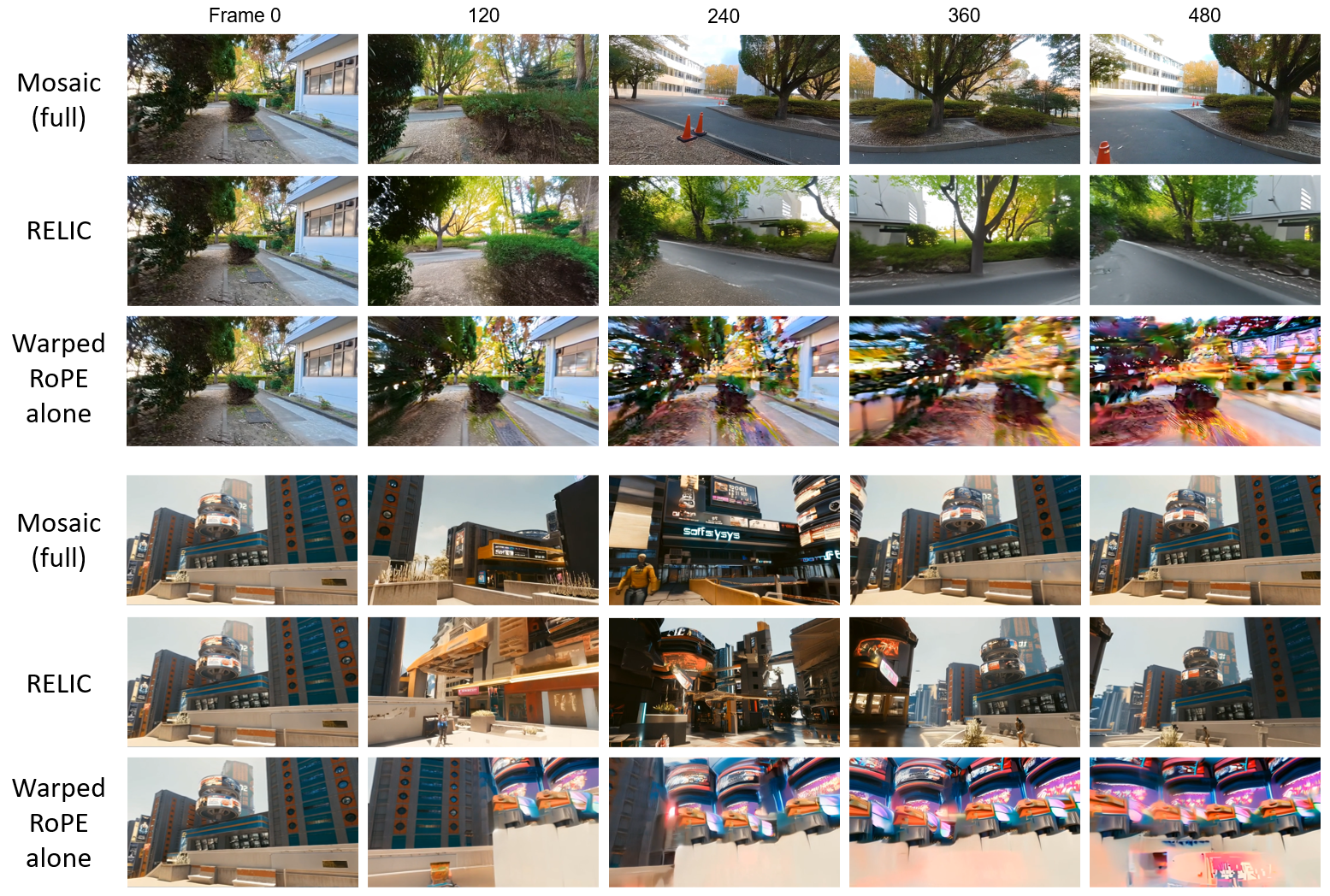}
    \vspace{-0.8em}
    \caption{Comparison of different autoregressive video generation systems. MosaicMem (full) maintains both visual quality and temporal consistency under few-step inference. RELIC’s memory mechanism introduces retrieval errors, while MosaicMem using only Warped RoPE may collapse in extreme cases.}
    \label{fig:camera}
    \vspace{-0.2em}
\end{figure}

\noindent\textbf{Results.} We present the results of memory manipulation in Fig.~\ref{fig:manipulation}. We can stitch multiple scenes together on a horizontal plane, enabling users to freely transition between different environments during exploration. For example, as shown in Fig.~\ref{fig:manipulation}(a), a user can move from a medieval scene to a modern street, explore the modern environment for a while, and then seamlessly return to the medieval scene. We can also create dreamlike scenes by flipping the Mosaic Memory and registering it in the sky, which is shown in Fig.~\ref{fig:manipulation}(b). This allows users to look up and observe another street appearing above them, forming a surreal Inception-like spatial connection between the ground scene and the scene in the sky.

\subsection{Mosaic Forcing}\label{sec:autoregressive}
In this section, we convert the originally bidirectional MosaicMem model into an autoregressive (AR) video generator using Causal Forcing \cite{zhu2026causal}, an upgraded version of Self Forcing \cite{huang2025self,chen2024diffusion}, and Rolling Forcing \cite{liu2025rolling}. We refer to this AR variant as Mosaic Forcing. Specifically, we distill the pretrained bidirectional MosaicMem diffusion model into a causal architecture following the Causal Forcing pipeline, and further incorporate the Rolling Forcing strategy during long-horizon generation to improve temporal consistency and reduce error accumulation. The resulting model achieves real-time generation at 16 FPS with a resolution of 640 $\times$ 360. We further compare Mosaic Forcing with existing real-time autoregressive video generation systems, including RELIC \cite{hong2025relic} and Matrix-Game 2.0 \cite{he2025matrix}. In addition to the metrics used previously, we follow the VBench protocol \cite{huang2024vbench}, evaluating Subject Consistency, Background Consistency, Motion Smoothness, Temporal Flickering, Aesthetic Quality, and Imaging Quality. The total quality score is computed as the average of these metrics.

The quantitative results and qualitative results are presented in Table \ref{tab:mosaic-forcing} and Figure \ref{fig:camera}, respectively. Owing to its precise camera-motion adherence and strong prompt-following capability, MosaicMem significantly outperforms the two baseline models across all metrics, particularly under large camera motions.

In addition, we observe another issue with Warped RoPE in certain extreme scenarios (e.g., when the camera motion is extremely slow): objects newly observed in previous frames tend to be repeatedly generated near the image boundaries. Introducing Warped Latent effectively resolves this issue.

\begin{table}[t]
\centering
\tiny
\setlength{\tabcolsep}{0.5pt}
\renewcommand{\arraystretch}{1.1}

\begin{tabular}{p{2.2cm}ccccccc cc cc}
\toprule

\textbf{Method}
& \multicolumn{7}{c}{\textbf{Quality Score ($\uparrow$)}}
& \multicolumn{2}{c}{\textbf{Consistency ($\uparrow$)}}
& \multicolumn{2}{c}{\textbf{Camera Control}} \\

\cmidrule(lr){2-8} \cmidrule(lr){9-10} \cmidrule(lr){11-12}

& Total
& \makecell{Subject\\Consist}
& \makecell{Bg\\Consist}
& \makecell{Motion\\Smooth}
& \makecell{Temporal\\Flicker}
& \makecell{Aesthetic\\Quality}
& \makecell{Imaging\\Quality}

& PSNR
& SSIM

& \makecell{RotErr ($^\circ$)\\$\downarrow$}
& \makecell{TransErr\\$\downarrow$}

\\

\midrule

Matrix-Game
& 75.11
& 82.40 & 87.92 & 88.35 & 89.10 & 43.12 & 59.77
& 18.57 & 0.524
& 5.32 & 0.38 \\

RELIC
& 79.08
& 86.21 & 91.08 & 94.12 & 92.05 & 47.01 & 64.02
& 20.23 & 0.591
& 4.99 & 0.36 \\

\midrule

MosaicMem-WRoPE
& 77.81
& 85.03 & 90.41 & 92.73 & 91.22 & 45.88 & 61.60
& 19.01 & 0.566
& 1.63 & 0.16 \\

MosaicMem (full)
& \textbf{81.11}
& \textbf{88.32} & \textbf{93.40} & \textbf{96.58} & \textbf{94.21} & \textbf{48.15} & \textbf{65.97}
& \textbf{21.57} & \textbf{0.652}
& \textbf{0.89} & \textbf{0.11} \\

\bottomrule
\end{tabular}
\vspace{0.4em}
\caption{Quantitative comparison on AR video generation, 
evaluating video quality and camera accuracy. 
MosaicMem achieves the best performance across all metrics.}
\label{tab:mosaic-forcing}
\vspace{-3.5em}
\end{table}

\section{Related work}

\textbf{Spatial Memory in Video Generation.}
Spatial memory methods can be categorized as explicit or implicit. Explicit approaches build persistent geometry, including point clouds \cite{ren2025gen3c,zhao2025spatia,wu2025videoworld} and surfels \cite{li2025vmem}: GEN3C \cite{ren2025gen3c} maintains a point-cloud cache, Spatia \cite{zhao2025spatia} updates memory via visual SLAM, and VMem \cite{li2025vmem} retrieves views using surfel-indexed visibility. While geometrically consistent, these methods struggle with complex dynamics. Implicit approaches \cite{worldlabs2025rtfm,xiao2025worldmem,oshima2025worldpack,yu2025contextasmem,sun2025worldplay} store past frames in learned representations, with WorldMem \cite{xiao2025worldmem} retrieving by FOV overlap, Context-as-memory \cite{yu2025contextasmem} applying rule-based selection, and WorldPack \cite{oshima2025worldpack} employed  trajectory packing. MosaicMem adopts a hybrid strategy, retrieving spatially aligned memory patches guided by 3D information while integrating them implicitly within the model.

\noindent\textbf{Camera-Controlled Video Generation.}
Explicit camera control is crucial for temporally coherent video generation \cite{yu2025prior,yu2025spiral}. Prior work conditions models on camera parameters directly \cite{wang2024motionctrl}, lifts 2D inputs to 3D and renders point clouds for view-consistent generation \cite{bahmani2025camtrol,ren2025gen3c,cao2025uni3c,feng2025i2vcontrol}, or encodes camera rays via pixel-wise Plücker embeddings \cite{xu2024camco,bahmani2024vd3d,kuang2024collaborative,he2025cameractrl,he2025cameractrl2,sitzmann2021light,yu2025egosim}. The recently proposed PE-Field \cite{bai2025pef} extends 2D positional encodings into a structured 3D field, enabling DiT to model geometry directly in 3D space for novel view synthesis. In parallel, our Warped RoPE adopts a similar geometric reprojection principle but extends it to video by incorporating temporal coordinates, allowing memory patches originating from different timesteps to be consistently aligned. We also adopt a complementary Warped Latent mechanism that directly transforms retrieved memory patches in latent feature space, improving autoregressive video generation. Moreover, we introduce PRoPE \cite{li2025prope} as a camera conditioning mechanism, capturing relative camera geometry via positional encoding and enabling  improved  controllability.

\vspace{-1.0em}
\section{Conclusion}
\vspace{-0.5em}
We present \emph{Mosaic Memory (MosaicMem)}, a \emph{patch-and-compose} spatial memory that bridges explicit and implicit paradigms for long-horizon, camera-controlled video generation. By lifting video patches for precise localization and retrieval while retaining attention-based conditioning for dynamic evolution, MosaicMem can retrieve accurate spatial memory, depict precise egomotion and enable rich prompt-driven dynamics. Combined with \emph{PRoPE} for finer viewpoint control and a new benchmark for revisits under dynamic changes, MosaicMem outperforms both paradigms and supports long navigation rollouts, direct memory manipulation for scene editing, and \emph{autoregressive rollout generation}.

\bibliographystyle{splncs04}
\bibliography{main}
\end{document}